# Models and Selection Criteria for Regression and Classification

**David Heckerman and Christopher Meek**
Microsoft Research
Redmond WA 98052-6399
heckerma@microsoft.com, meek@microsoft.com

## Abstract

When performing regression or classification, we are interested in the conditional probability distribution for an *outcome* or *class* variable $Y$ given a set of *explanatory* or *input* variables $\mathbf{X}$. We consider Bayesian models for this task. In particular, we examine a special class of models, which we call Bayesian regression/classification (BRC) models, that can be factored into independent conditional $(y|\mathbf{x})$ and input $(\mathbf{x})$ models. These models are convenient, because the conditional model (the portion of the full model that we care about) can be analyzed by itself. We examine the practice of transforming arbitrary Bayesian models to BRC models, and argue that this practice is often inappropriate because it ignores prior knowledge that may be important for learning. In addition, we examine Bayesian methods for learning models from data. We discuss two criteria for Bayesian model selection that are appropriate for repression/classification: one described by Spiegelhalter et al. (1993), and another by Buntine (1993). We contrast these two criteria using the prequential framework of Dawid (1984), and give sufficient conditions under which the criteria agree.

## 1 Introduction

Most work on learning Bayesian networks from data has concentrated on the determination of relationships among a set of variables. This task, which we call *joint analysis*[1], has applications in causal discovery and the prediction of a set of observations. Another important task is *regression/classification*: the determination of a conditional probability distribution for an *outcome* or *class* variable $Y$ given a set of *explanatory* or *input* variables $\mathbf{X}$. When $Y$ has a finite number of states we refer to the task as *classification*. Otherwise we refer to the task as *regression*.

In this paper, we examine parametric models for the regression/classification task. In Section 2, we examine a special class of models, which we call Bayesian regression/classification (BRC) models, that can be factored into independent conditional $(y|\mathbf{x})$ and input $(\mathbf{x})$ models. These models are convenient, because the conditional model (the portion of the full model that we care about) can be analyzed alone. In Section 3, we examine the practice of transforming arbitrary Bayesian models to BRC models, and argue that this practice is often inappropriate because it ignores prior knowledge that may be important for learning.

Also in this paper, we discuss Bayesian methods for learning models from data. In Section 4, we compare Bayesian model averaging and model selection. In Section 5, we discuss two criteria for Bayesian model selection that are appropriate for regression/classification: one described by Spiegelhalter et al. (1993), and another by Buntine (1993). We contrast these two criteria using the prequential framework of Dawid (1984), and give sufficient conditions under which the criteria agree.

The terminology and notation we need is as follows. We denote a variable by an upper-case letter (e.g., $X, Y, X_i, \Theta$), and the state or value of a corresponding variable by that same letter in lower case (e.g., $x, y, x_i, \theta$). We denote a set of variables by a bold-face upper-case letter (e.g., $\mathbf{X}, \mathbf{Y}, \mathbf{X}_i$). We use a corresponding bold-face lower-case letter (e.g., $\mathbf{x}, \mathbf{y}, \mathbf{x}_i$) to denote an assignment of state or value to each vari-able in a given set. We say that variable set $\mathbf{X}$ is in *configuration* $\mathbf{x}$. We use $p(\mathbf{X} = \mathbf{x}|\mathbf{Y} = \mathbf{y})$ (or $p(\mathbf{x}|\mathbf{y})$ as a shorthand) to denote the probability or probability density that $\mathbf{X} = \mathbf{x}$ given $\mathbf{Y} = \mathbf{y}$. We also use $p(\mathbf{x}|\mathbf{y})$ to denote the probability distribution (both mass functions and density functions) for $\mathbf{X}$ given $\mathbf{Y} = \mathbf{y}$. Whether $p(\mathbf{x}|\mathbf{y})$ refers to a probability, a probability density, or a probability distribution will be clear from context.

---
[1] This task is sometimes called *density estimation*.

We use $\mathbf{m}$ and $\theta_m$ to denote the structure and parameters of a model, respectively. When $(\mathbf{m}, \theta_m)$ is a Bayesian network for variables $\mathbf{Z}$, we write the usual factorization as

$$p(z_1, \ldots, z_N | \theta_m, \mathbf{m}) = \prod_{i=1}^{N} p(z_i | \mathbf{pa}_i, \theta_m, \mathbf{m}) \quad (1)$$

where $\mathbf{Pa}_i$ are the variables corresponding to the parents of $Z_i$ in $\mathbf{m}$. We refer to $p(z_i|\mathbf{pa}_i, \theta_m, \mathbf{m})$ as the *local distribution function* for $Z_i$.

## 2 Models for Regression/Classification

In this section, we examine various parametric models for the task of regression/classification. Models for this task are of two main types: conditional models and joint models. A *conditional model* is of the form $p(y|\mathbf{x}, \theta_m, \mathbf{m})$. A *joint model* is of the form $p(y, \mathbf{x}|\theta_m, \mathbf{m})$. We use a joint model for regression/classification by performing probabilistic inference to obtain $p(y|\mathbf{x}, \theta_m, \mathbf{m})$.

Examples of joint models include Bayesian networks. Figure 1a shows the structure of a *naive Bayes* model in which the variables $\mathbf{X}$ are mutually independent given $Y$. Suppose $Y$ has $r$ states $y^1, \ldots, y^r$, each $X_i$ is binary with states $x_i^1$ and $x_i^2$, and each local distribution function is a collection of multinomial distributions (one distribution for each parent configuration). For this example, it is not difficult to derive the corresponding conditional model (see, for example, Bishop, 1995, Chapter 6). Namely, we have

$$\lambda_{k\mathbf{x}} \equiv \log \frac{p(y^k|\mathbf{x}, \theta_m, \mathbf{m})}{p(y^1|\mathbf{x}, \theta_m, \mathbf{m})} = \log \frac{\theta(y^k)}{\theta(y^1)} + \sum_{i=1}^{n} \log \frac{\theta(x_i|y^k)}{\theta(x_i|y^1)} \quad (2)$$

for $k = 2, \ldots, r$. After some algebra, Equation 2 becomes

$$\lambda_{k\mathbf{x}} = \left( \log \frac{\theta(y^2)}{\theta(y^1)} + \sum_{i=1}^{n} \log \frac{\theta(x_i^1|y^2)}{\theta(x_i^1|y^1)} \right) \quad (3)$$
$$+ \sum_{i=1}^{n} I(x_i^2) \left( \frac{\theta(x_i^2|y^2)}{\theta(x_i^2|y^1)} - \frac{\theta(x_i^1|y^2)}{\theta(x_i^1|y^1)} \right)$$

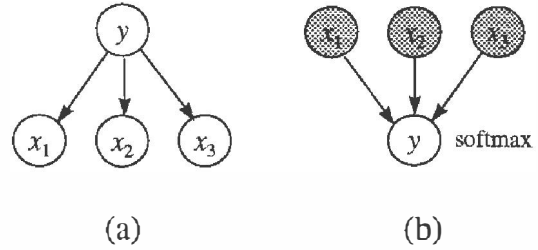

Figure 1: (a) A naive Bayes model for classification. (b) A linear softmax regression model that has the same conditional distribution for $Y$.

where $I(x_i^2)$ is the indicator variable that is equal to 1 if and only if $x_i = x_i^2$. Consequently, we have

$$p(y^k|\mathbf{x}, \theta_m, \mathbf{m}) = \frac{e^{\lambda_{k\mathbf{x}}}}{1 + \sum_{j=2}^{r} e^{\lambda_{j\mathbf{x}}}} \quad (4)$$
$$\equiv \text{softmax}(\lambda_{1\mathbf{x}}, \ldots, \lambda_{k\mathbf{x}})$$

where each $\lambda_{k\mathbf{x}}$ is a linear function of $I(x_1^2), \ldots, I(x_n^2)$.

This conditional model $p(y|\mathbf{x}, \theta_m, \mathbf{m})$ is a type of generalized linear model known as a *linear softmax regression*.[2] We can display the structure of this conditional model as a Bayesian network, as shown in Figure 1b. In the figure, the input nodes $\mathbf{X}$ are shaded to indicate that we observe them and hence do not care about their joint distribution.

Now let us specialize our discussion to Bayesian models for regression/classification. In the Bayesian approach, we encode our uncertainty about $\theta_m$ and $\mathbf{m}$ using probability distributions $p(\theta_m|\mathbf{m})$ and $p(\mathbf{m})$, respectively. Thus, the Bayesian variant of a joint model takes the form

$$p(y, \mathbf{x}, \theta_m, \mathbf{m}) = p(\mathbf{m}) \, p(\theta_m|\mathbf{m}) \, p(y, \mathbf{x}|\theta_m, \mathbf{m}) \quad (5)$$

We refer to this model as a *Bayesian joint* (BJ) model.

We define a Bayesian analogue to a conditional model as follows. Suppose that $\theta_m$ can be decomposed into parameters $(\theta_\mathbf{x}, \theta_{y|\mathbf{x}})$ such that

$$p(y, \mathbf{x}|\theta_m, \mathbf{m}) = p(\mathbf{x}|\theta_\mathbf{x}, \mathbf{m}) \, p(y|\mathbf{x}, \theta_{y|\mathbf{x}}, \mathbf{m}) \quad (6)$$

$$p(\theta_m|\mathbf{m}) = p(\theta_\mathbf{x}|\mathbf{m}) \, p(\theta_{y|\mathbf{x}}|\mathbf{m}) \quad (7)$$

In this case, given data $D = ((y_1, \mathbf{x}_1), \ldots, (y_N, \mathbf{x}_N))$, assumed to be a random sample from the true distribution of $Y$ and $\mathbf{X}$, we have

$$p(\theta_m|y, \mathbf{x}, \mathbf{m}) = \{p(\mathbf{x}|\theta_\mathbf{x}, \mathbf{m}) p(\theta_\mathbf{x}|\mathbf{m})\}$$
$$\cdot \{p(y|\mathbf{x}, \theta_{y|\mathbf{x}}, \mathbf{m}) p(\theta_{y|\mathbf{x}}|\mathbf{m})\}$$

---

[2]Although $Y$ has a finite number of states, this model is commonly referred to as a regression.

Consequently, we can analyze the marginal ($\mathbf{x}$) and conditional ($y|\mathbf{x}$) terms independently. In particular, if we care only about the conditional distribution, we can analyze it on its own. We call this model defined by Equations 6 and 7 a *Bayesian regression/classification* (BRC) model. Simple examples of BRC models include ordinary linear regression (e.g., Gelman et al., 1995, Chapter 8), and generalized linear models (e.g., Bishop, 1995, Chapter 10).

Note that our Bayesian analogue to the conditional model is a special case of a BJ model. One could imagine using a Bayesian model that encodes only the conditional likelihood $p(y|\mathbf{x}, \boldsymbol{\theta}_m, \mathbf{m})$ and a joint distribution for $\boldsymbol{\theta}_m$ and $\mathbf{m}$. However, this approach is flawed, because it may miss important relationships among the domain variables or their parameters that are important for learning. In the following section, we consider an example of this point.

## 3  Embedded Regression/Classification Models

A special class of BRC models is suggested by the following observation. For many BJ models, the conditional likelihood $p(y|\mathbf{x}, \boldsymbol{\theta}_m, \mathbf{m})$ is a simple function of $\mathbf{x}$, whereas the expression for the input likelihood $p(\mathbf{x}|\boldsymbol{\theta}_m, \mathbf{m})$ is more complicated. For example, given a naive-Bayes model in which the variables $\mathbf{X}$ are mutually independent given $Y$, the conditional likelihood is a simple generalized linear model, but the input likelihood is a mixture distribution. Thus, assuming we are interested in the task of regression/classification, we can imagine extracting the conditional likelihood from a BJ model, and embedding it in a BRC model. In particular, given a BJ model $(\boldsymbol{\theta}_m, \mathbf{m})$, we can create a BRC model $(\boldsymbol{\theta}'_m, \mathbf{m}')$ in which $p(y|\mathbf{x}, \boldsymbol{\theta}'_m, \mathbf{m}') = p(y|\mathbf{x}, \boldsymbol{\theta}_m, \mathbf{m})$. We say that $(\boldsymbol{\theta}'_m, \mathbf{m}')$ is a *Bayesian embedded regression/classification* (BERC) *model obtained from* $(\boldsymbol{\theta}'_m, \mathbf{m}')$.

Several researchers have suggested using BERC models, at least implicitly (see Bishop, 1995, Chapter 10, and references therein). An example of a BERC model obtained from a naive Bayes model is shown in Figure 2. If a BERC model $(\boldsymbol{\theta}'_m, \mathbf{m}')$ is obtained from a model which is itself a BERC model, we refer to $(\boldsymbol{\theta}'_m, \mathbf{m}')$ as a trivial BERC model. The BERC model is Figure 2 is non-trivial.

For any Bayesian network with finite-state variables, it is not difficult to obtain its corresponding BERC model. Let $X_1, \ldots, X_{n_h}, Y, X_{n_h+1}, \ldots, X_n$ be a total ordering on the variables that is consistent with $\mathbf{m}$, such that $Y$ appears as late as possible in the ordering. The latter condition says that the node corresponding to $Y$ is an ancestor of each of the nodes corresponding to $X_{n_h+1}, \ldots, X_n$. Given this ordering, we can factor the joint distribution for $Y, X_1, \ldots, X_n$ as follows:

$$p(y, \mathbf{x}|\boldsymbol{\theta}_m, \mathbf{m}) = \left( \prod_{i=1}^{n_h} p(x_i|\mathbf{pa}_i, \boldsymbol{\theta}_m, \mathbf{m}) \right)$$
$$\cdot p(y|\mathbf{pa}_y, \boldsymbol{\theta}_m, \mathbf{m}) \cdot \left( \prod_{i=n_h+1}^{n} p(x_i|\mathbf{pa}_i, \boldsymbol{\theta}_m, \mathbf{m}) \right)$$

where $Y$ does not appear in any parent set $\mathbf{Pa}_i$ in the first product. Normalizing to obtain $p(y|\mathbf{x}, \boldsymbol{\theta}_m, \mathbf{m})$, taking a ratio, and canceling like terms, we obtain

$$\lambda_{k\mathbf{x}} = \log \frac{\theta(y^k|\mathbf{pa}_y)}{\theta(y^1|\mathbf{pa}_y)} + \sum_{i=n_h+1}^{n} \log \frac{\theta(x_i|\mathbf{pa}_i^k)}{\theta(x_i|\mathbf{pa}_i^1)} \quad (8)$$

where $\mathbf{pa}_i^k$ is a configuration of $\mathbf{Pa}_i$ in which $y = y^k$, $k = 1, \ldots, r$. (Depending on $\mathbf{m}$, some of the terms in the sum may cancel as well.) We can trivially rewrite Equation 8 as

$$\lambda_{k\mathbf{x}} = I(x_1) \cdots I(x_n)$$
$$\cdot \left\{ \log \frac{\theta(y^k|\mathbf{pa}_y)}{\theta(y^1|\mathbf{pa}_y)} + \sum_{i=n_h+1}^{n} \log \frac{\theta(x_i|\mathbf{pa}_i^k)}{\theta(x_i|\mathbf{pa}_i^1)} \right\} \quad (9)$$

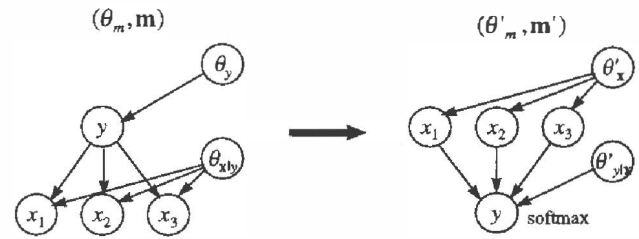

Figure 2: A BERC model obtained from a naive-Bayes model.

Equation 9 shows that an BERC model is a *polynomial softmax regression on the indicator variables* $I(x_1), \ldots, I(x_n)$. Note that there are polynomial softmax regressions that cannot be obtained from any Bayesian network.

Although BERC models are convenient, we find non-trivial BERC models to be problematic. In particular, consider a BERC model $(\boldsymbol{\theta}'_m, \mathbf{m}')$ obtained from a non-BERC model $(\boldsymbol{\theta}_m, \mathbf{m})$. Whereas in the BERC model, observations of $\mathbf{X}$ are necessarily uninformative about $\boldsymbol{\theta}_{y|\mathbf{x}}$, such observations may be informative in the original model $(\boldsymbol{\theta}_m, \mathbf{m})$. Thus, in constructing the BERC model, we may be ignoring parts of our prior knowledge that are important for learning.

To illustrate this point, consider the naive-Bayes model for binary variables $Y, X_1, X_2, X_3$. The mapping from $\boldsymbol{\theta}_m$ to $\boldsymbol{\theta}_\mathbf{x}$ is shown in Figure 3. It is not

$$\theta(x_1^1 x_2^1 x_3^1) = \theta(y^1)\,\theta(x_1^1|y^1)\,\theta(x_2^1|y^1)\,\theta(x_3^1|y^1) + \theta(y^2)\,\theta(x_1^1|y^2)\,\theta(x_2^1|y^2)\,\theta(x_3^1|y^2)$$

$$\theta(x_1^1 x_2^1 x_3^2) = \theta(y^1)\,\theta(x_1^1|y^1)\,\theta(x_2^1|y^1)(1-\theta(x_3^1|y^1)) + \theta(y^2)\,\theta(x_1^1|y^2)\,\theta(x_2^1|y^2)(1-\theta(x_3^1|y^2))$$

$$\theta(x_1^1 x_2^2 x_3^1) = \theta(y^1)\,\theta(x_1^1|y^1)\,\theta(x_2^1|y^1)\,\theta(x_3^1|y^1) + \theta(y^2)\,\theta(x_1^1|y^2)\,\theta(x_2^1|y^2)\,\theta(x_3^1|y^2)$$

$$\theta(x_1^1 x_2^2 x_3^2) = \theta(y^1)\,\theta(x_1^1|y^1)(1-\theta(x_2^1|y^1))(1-\theta(x_3^1|y^1)) + \theta(y^2)\,\theta(x_1^1|y^2)(1-\theta(x_2^1|y^2))(1-\theta(x_3^1|y^2))$$

$$\theta(x_1^2 x_2^1 x_3^1) = \theta(y^1)(1-\theta(x_1^1|y^1))\,\theta(x_2^1|y^1)\,\theta(x_3^1|y^1) + \theta(y^2)(1-\theta(x_1^1|y^2))\,\theta(x_2^1|y^2)\,\theta(x_3^1|y^2)$$

$$\theta(x_1^2 x_2^1 x_3^2) = \theta(y^1)(1-\theta(x_1^1|y^1))\,\theta(x_2^1|y^1)(1-\theta(x_3^1|y^1)) + \theta(y^2)(1-\theta(x_1^1|y^2))\,\theta(x_2^1|y^2)(1-\theta(x_3^1|y^2))$$

$$\theta(x_1^2 x_2^2 x_3^1) = \theta(y^1)(1-\theta(x_1^1|y^1))\,\theta(x_2^1|y^1)\,\theta(x_3^1|y^1) + \theta(y^2)(1-\theta(x_1^1|y^2))\,\theta(x_2^1|y^2)\,\theta(x_3^1|y^2)$$

Figure 3: The mapping from $\theta_m$ to $\theta_x$ for the naive-Bayes model where $Y$ renders $X_1, X_2$, and $X_3$ mutually independent. We use $\theta(y)$, $\theta(x_1,x_2,x_3)$, and $\theta(x_i|y)$ to denote $p(y|\theta_m,\mathbf{m})$, $p(x_1,x_2,x_3|\theta_m,\mathbf{m})$, and $p(x_i|y,\theta_m,\mathbf{m})$, respectively.

difficult to show that the rank of the Jacobian matrix $\partial\theta_x/\partial\theta_m$ is full (i.e., equal to the number of non-redundant parameters in $\theta_m$) for almost all values of $\Theta_m$ (see, e.g., Geiger et al., 1996). It follows that, for almost every point $\theta_m^*$ in $\theta_m$, there is an inverse mapping from $\theta_x$ to $\theta_m$ in a neighborhood around $\theta_m^*$.[3] Consequently, the possible values that $\theta_m$ (and hence $\theta_{y|x}$) can assume will depend on the value of $\theta_x$, and observations of $\mathbf{X}$ will inform $\theta_{y|x}$ through $\theta_x$.

In general, given two variables (random or otherwise) $A$ and $B$, if the possible values that can be assumed by $A$ depend on the value of $B$, then $A$ is said to be *variationally dependent* on $B$. In our example, $\Theta_{y|x}$ is variationally dependent on $\Theta_x$. Such variational dependence is not limited to this example. For any model $(\theta_m, \mathbf{m})$, if the rank of the Jacobian matrix for the mapping from $\theta_m$ to $\theta_x$ is full, then $\Theta_m$ (and hence $\Theta_{y|x}$) is variationally dependent on $\theta_x$. Geiger et al. (1996) conjecture that, for naive-Bayes models in which all variables are binary, the rank of the Jacobian matrix for the mapping from $\theta_m$ to $\theta_x$ is almost everywhere full. In addition, Goodman (1974) and Geiger et al. (1996) could identify only one naive-Bayes model in which the Jacobian matrix was not of full rank almost everywhere. Thus, the use of non-trivial BERC models—at least those obtained from most naive Bayes models—is suspect.

Note that our remarks extend to non-Bayesian analyses. For example, in a classical analysis, a polynomial softmax regression should not be substituted for a Bayesian network. In the former model, $\Theta_{y|x}$ and $\Theta_x$ are variationally independent. In the latter model, $\Theta_{y|x}$ and $\Theta_x$ are variationally dependent, and observations of $\mathbf{X}$ often will influence the estimate of $\theta_{y|x}$. More generally, conditional models—often referred to as regression/classification models—should not be used without consideration of variational dependencies that may arise from the joint model.

## 4 Learning Regression/Classification Models: Averaging Versus Selection

Now that we have examined several classes of models for the regression/classification task, let us concentrate on Bayesian methods for learning such models.

First, consider *model averaging*. Given a random sample $D$ from the true distribution of $Y$ and $\mathbf{X}$, we compute the posterior distributions for each $\mathbf{m}$ and $\theta_m$ using Bayes' rule:

$$p(\mathbf{m}|D) = \frac{p(\mathbf{m})\,p(D|\mathbf{m})}{\sum_{\mathbf{m}'} p(\mathbf{m}')\,p(D|\mathbf{m}')}$$

$$p(\theta_m|D,\mathbf{m}) = \frac{p(\theta_m|\mathbf{m})\,p(D|\theta_m,\mathbf{m})}{p(D|\mathbf{m})}$$

where

$$p(D|\mathbf{m}) = \int p(D|\theta_m,\mathbf{m})\,p(\theta_m|\mathbf{m})\,d\theta_m$$

With these quantities in hand, we can determine the conditional distribution for $Y$ given $\mathbf{X}$ in the next case to be seen by averaging over all possible model structures and their parameters:

$$p(y|\mathbf{x},D) = \sum_m p(\mathbf{m}|D)\,p(y|\mathbf{x},D,\mathbf{m}) \qquad (10)$$

$$p(y|\mathbf{x},D,\mathbf{m}) = \int p(y|\mathbf{x},\theta_m,\mathbf{m})\,p(\theta_m|D,\mathbf{m})d\theta_m \qquad (11)$$

---
[3]The parameters $\theta_m$ are said to be *locally identifiable* given observations of $\mathbf{X}$ (e.g., Goodman, 1974).

Note that joint analysis is handled in essentially the same way. For example, to determine the joint distribution of $Y$ and $\mathbf{X}$ in the next case to be seen, we use

$$p(y,\mathbf{x}|D) = \sum_m p(\mathbf{m}|D) p(y,\mathbf{x}|D,\mathbf{m}) \quad (12)$$

$$p(y,\mathbf{x}|D,m) = \int p(y,\mathbf{x}|\theta_m,\mathbf{m})\ p(\theta_m|D,\mathbf{m}) d\theta_m \quad (13)$$

Model averaging, however, is not always appropriate for an analysis. For example, only one or a few models may be desired for domain understanding or for fast prediction. In these situations, we select one or a few "good" model structures from among all possible models, and use them as if they were exhaustive. This procedure is known as *model selection* when one model is chosen, and *selective model averaging* when more than one model is chosen. Of course, model selection and selective model averaging are also useful when it is impractical to average over all possible model structures.

When our goal is model selection, a "good" model for joint analysis may not be a good model for regression/classification, and vice versa. Scores that define "good" model structures are commonly known as *criteria*. A criterion commonly used for joint analysis is the logarithm of the relative posterior probability of the model structure $\log p(\mathbf{m}, D) = \log p(\mathbf{m}) + \log p(D|\mathbf{m})$. This criterion is *global* in the sense that it is equally sensitive to possible dependencies among all variables. Criteria for regression/classification, should be *local* in the sense that they concentrate on how well $\mathbf{X}$ classifies $Y$. In the following section, we examine two such criteria.

## 5   Prequential Criteria for Regression/Classification

The criteria that we discuss can be understood in terms of Dawid's (1984) predictive sequential or *prequential* method. A simple example of this method, applied to joint analysis, yields the posterior-probability criterion. Let us consider this example first.

To simplify the discussion, let us assume that that $p(\mathbf{m})$ is uniform, so that the joint-analysis criterion reduces to the log-marginal-likelihood $\log p(D|\mathbf{m})$.[4] From the chain rule of probability, the log marginal likelihood is given by

$$\log p(D|\mathbf{m}) = \sum_{l=1}^{N} \log p(y_l,\mathbf{x}_l|y_1,\mathbf{x}_1,\ldots,y_{l-1},\mathbf{x}_{l-1},\mathbf{m})$$

[4] The generalization to non-uniform model priors is straightforward.

The term $p(y_l,\mathbf{x}_l|y_1,\mathbf{x}_1,\ldots,y_{l-1},\mathbf{x}_{l-1},\mathbf{m})$ is the prediction for $(y_l,\mathbf{x}_l)$ made by model structure $\mathbf{m}$ after averaging over its parameters (Equation 13). The log of this term can be thought of as the utility for this prediction.[5] Thus, a model structure with the highest log marginal likelihood is also a model structure that is the best sequential predictor of the data $D$ given the logarithmic utility function.

Let us now consider local criteria that are more appropriate for the task of regression/classification. To keep the discussion brief, we discuss only the logarithmic utility function, although other utility functions may be more reasonable for a given problem. At least two prequential criteria are reasonable. In one situation, we imagine that we see pairs $(y_l,\mathbf{x}_l)$ sequentially. As a result, we obtain a criterion that Spiegelhalter et al. (1993) call a *conditional node monitor*:

$$\mathrm{CNM}(D,\mathbf{m}) = \sum_{l=1}^{N} \log p(y_l|\mathbf{x}_l, y_1,\mathbf{x}_1,\ldots,y_{l-1},\mathbf{x}_{l-1},\mathbf{m}) \quad (14)$$

In another situation, we imagine that we first see all of the input data $\mathbf{x}_1,\ldots,\mathbf{x}_N$, and then see the class data sequentially. Consequently, we obtain the following *class sequential criterion*:

$$\mathrm{CSC}(D,\mathbf{m}) = \sum_{l=1}^{N} \log p(y_l|y_1,\ldots,y_{l-1},\mathbf{x}_1,\ldots,\mathbf{x}_N,\mathbf{m}) \quad (15)$$

Buntine (1993) used this criterion for selection among decision-tree structures.

Spiegelhalter et al. (1993) describe a set of assumptions—essentially, parameter independence and Dirichlet priors—under with the conditional node monitor can be computed efficiently in closed form. Under these same assumptions, the exact computation of the class sequential criterion is exponential in the sample size $N$. Monte-Carlo or asymptotic techniques can be used to perform the computation for large $N$ (see, e.g., Heckerman, 1995).

We have applied both criteria to small Bayesian networks and small data sets chosen arbitrarily. In all cases, we have found that the two criteria differ. Nonetheless, there are conditions under which the two criteria are the same. In particular, we can rewrite the two criteria as follows:

$$\mathrm{CNM}(D,\mathbf{m}) = \sum_{l=1}^{N} \log \frac{p(y_l,\mathbf{x}_l|y_1,\mathbf{x}_1,\ldots,y_{l-1},\mathbf{x}_{l-1},\mathbf{m})}{p(\mathbf{x}_l|y_1,\mathbf{x}_1,\ldots,y_{l-1},\mathbf{x}_{l-1},\mathbf{m})} \quad (16)$$

[5] The utility $\log x$ is also known as a *scoring rule*. Bernardo (1979) shows that this scoring rule has several desirable properties.

$$\text{CSC}(D, \mathbf{m}) = \log \frac{p(y_1, \ldots, y_N, \mathbf{x}_1, \ldots, \mathbf{x}_N | \mathbf{m})}{p(\mathbf{x}_1, \ldots, \mathbf{x}_N | \mathbf{m})} \quad (17)$$

$$= \sum_{l=1}^{N} \log \frac{p(y_l, \mathbf{x}_l | y_1, \mathbf{x}_1, \ldots, y_{l-1}, \mathbf{x}_{l-1}, \mathbf{m})}{p(\mathbf{x}_l | \mathbf{x}_1, \ldots, \mathbf{x}_{l-1}, \mathbf{m})}$$

Therefore, the two criteria will agree when

$$p(\mathbf{x}_l | y_1, \mathbf{x}_1, \ldots, y_{l-1}, \mathbf{x}_{l-1}, \mathbf{m}) = p(\mathbf{x}_l | \mathbf{x}_1, \ldots, \mathbf{x}_{l-1}, \mathbf{m}) \quad (18)$$

for $l = 0, \ldots, N-1$. It is not difficult to show that Equation 18 holds whenever $(\theta_m, \mathbf{m})$ is a BRC model. Thus, the two criteria agree for BRC models.

## 6 Discussion

Several researchers have demonstrated that Bayesian networks for both the joint analysis and regression/classification tasks provide better predictions when local distribution functions are encoded with a small number of parameters, as is the case with the use of decision trees, decision graphs, and causal-independence models (e.g., Friedman and Goldszmidt, 1996; Chickering et al., 1997; Meek and Heckerman, 1997). Despite our theoretical objections to the use of BERC models, they offer another parsimonious parameterization of local distribution functions, and may lead to better predictions in practice. For example, polynomial softmax regressions may be useful when a node and its parents are discrete. Experiments are needed to investigate these possibilities.

## Acknowledgments

We thank Max Chickering for useful discussions.